\pdfoutput=1
\documentclass[11pt]{article}

\usepackage[]{acl} %review
\usepackage{times}
\usepackage{latexsym}

\usepackage[T1]{fontenc}
\usepackage[utf8]{inputenc}

\usepackage{microtype}
\usepackage{graphicx}

\usepackage{inconsolata}
\usepackage{booktabs}
\usepackage{xcolor} % Added for color commands
\usepackage{multicol}
\usepackage{multirow}
\usepackage{subcaption}

\title{Language as a Label: Zero-Shot Multimodal Classification of Everyday Postures under Data Scarcity}
\author{MingZe Tang, Jubal Chandy Jacob\\
  Department of Computing Science \\
  University of Aberdeen \\
  \texttt{\{m.tang.24, j.chandyjacob.24\}@abdn.ac.uk} \\}

\begin{document}
\maketitle
\begin{abstract}
Recent Vision-Language Models (VLMs) enable zero-shot classification by aligning images and text in a shared space, a promising approach for data-scarce conditions. However, the influence of prompt design on recognizing visually similar categories, such as human postures, is not well understood. This study investigates how prompt specificity affects the zero-shot classification of sitting, standing, and walking/running on a small, 285-image COCO-derived dataset. A suite of modern VLMs, including OpenCLIP, MetaCLIP 2, and SigLip, were evaluated using a three-tiered prompt design that systematically increases linguistic detail. Our findings reveal a compelling, counter-intuitive trend: for the highest-performing models (MetaCLIP 2 and OpenCLIP), the simplest, most basic prompts consistently achieve the best results. Adding descriptive detail significantly degrades performance for instance, MetaCLIP 2's multi-class accuracy drops from 68.8\% to 55.1\% a phenomenon we term "prompt overfitting". Conversely, the lower-performing SigLip model shows improved classification on ambiguous classes when given more descriptive, body-cue-based prompts.
\end{abstract}

\section{Introduction}
Label scarcity is a central barrier for practical human action recognition from still images \cite{Wu2022EndtoEndZH}. Many deployments cannot acquire balanced annotations or run task specific training. Vision and language encoders mitigate this limitation by learning a shared embedding space in which text can serve as a label at inference time \cite{Radford2021LearningTV}. This paper studies whether careful wording of those text labels improves zero shot classification under data scarcity.

The task focuses on three everyday postures in still images, namely sitting, standing, and walking or running, using a small subset derived from COCO \cite{Lin2014MicrosoftCCA} with 230 images. Image content, preprocessing, and scoring are held fixed, and language acts as the only supervision at inference. Each image is embedded once at the native input size of the model and is scored by cosine similarity against one prompt per class. Prompt specificity is the sole experimental factor and follows a three tier design. Tier one uses a minimal label template. Tier two adds a short action cue. Tier three adds compact pose geometry that specifies body configuration. Prompts exclude scene, identity, and clothing terms so that differences arise only from pose description.

Evaluation covers multimodal encoders that align images and text, namely OpenCLIP, MetaCLIP and SigLIP. Vision only baselines include DINOv3 and a standard Vision Transformer paired with frozen sentence embeddings to form a heuristic zero shot classifier. A pose based baseline uses YOLOv11 Pose for key-point estimation together with a simple geometric decision rule. Results are reported as accuracy and macro F1 for each tier and each model, and qualitative analysis with gradient based visualisations assesses whether greater prompt specificity shifts attention toward pose relevant regions. The study provides an empirical protocol for zero shot recognition under data scarcity and a controlled comparison of prompt wording across modern encoders and non linguistic baselines.

\section{Related Works}

\subsection{Vision Language Models for Zero-Shot Classification}
At the core of modern zero-shot classification is the contrastive language image pre-training paradigm introduced by OpenCLIP, which aligns visual and textual representations in a shared embedding space using large collections of image–text pairs \cite{Radford2021LearningTV,Jia2021ScalingUV}. The objective draws paired images and texts closer while separating mismatched pairs, thereby encoding vision language correspondences \cite{Zhang2023VisionLanguageMF}. The resulting zero-shot classification mechanism is straightforward: given an input image and a set of class names, the model computes cosine similarity between the normalized image embedding and the text embeddings of natural language prompts such as “a photo of a \texttt{[class]}” with the highest similarity determines the predicted class \cite{Ghiasvand2025FewShotAL,Ghiasvand2025pFedMMAPF}.

OpenCLIP, reports competitive results on more than thirty vision datasets that range from optical character recognition to fine grained object classification, in some cases approaching fully supervised baselines without task specific tuning \cite{Radford2021LearningTV}. The same formulation has been adapted beyond image classification to object detection, semantic segmentation, video action recognition, and depth estimation \cite{Xu2023BenchmarkingZR,Zhou2022DetectingTC,Xu2022GroupViTSS}. More recent models such as SigLIP refine the approach, and dual encoder architectures are now common because they scale well and remain flexible across tasks \cite{Volkov2025ImageRW, Zhang2023VisionLanguageMF}.

\subsection{Low-Resource \& Low-Compute Image Understanding}
The need to balance performance and efficiency has motivated training free methods for resource constrained settings where computationally intensive training is not feasible \cite{Zhang2024DualMN}. These methods aim to extract more information from test samples and class names without updating parameters, which is helpful when both compute and labelled data are limited \cite{Zhang2024DualMN}. Parameter efficient adaptation is commonly used when full fine tuning is costly. Parameter efficient adaptation offers a middle ground. Techniques such as prompt tuning allow models to adapt to downstream tasks while most parameters remain fixed \cite{Mistretta2024ImprovingZG,Lester2021ThePO}. In addition, training free few shot methods such as TIP Adapter use cache models with support samples to encode few shot knowledge and combine it with zero-shot knowledge from text prompts through weighted integration \cite{Esbri2024MIVisionShotFA}.

Efficiency concerns also reflect domain specific factors. Direct zero-shot application can be affected by domain shift and by the inference cost of large models, especially when moving from general domain pre-training to specialized areas such as medical imaging \cite{Wang2025PretrainedVM,Liu2023VisualIT}. Video understanding adds further demand since many pipelines alternate learning of spatial and temporal information and therefore require substantial compute \cite{Bosetti2024TextEnhancedZA,Shao2020TemporalIN}. Together, these considerations suggest that vision and language models can be useful in specialized domains and low resource scenarios where fully supervised methods face data scarcity. In such cases, they can serve as a data multiplier by leveraging pre-trained knowledge when conventional approaches have little labelled data \cite{Zhang2024DualMN,Volkov2025ImageRW}.

\subsection{Multilinguality and Prompt Semantics for Posture Cues}
Early work on domain-specific zero-shot recognition represented actions through manually defined semantic concepts and attributes. This research demonstrated that carefully designed semantic encoding could enable recognition of action categories not seen during training \cite{Bosetti2024TextEnhancedZA,Zellers2017ZeroShotAR}. A parallel line of work replaced hand-specified concepts with distributional word embeddings of action names. More recent studies have extended this idea by treating the language modality itself as the main source of generalisation to new tasks and categories in video-oriented settings \cite{Bosetti2024TextEnhancedZA,Shao2020TemporalIN}.

Prompt design tailored to specific domains has since emerged as an important factor in improving performance on specialised tasks. Comparative studies between prompts derived directly from raw action labels and prompts generated by large language models show that the choice of description has a measurable effect on zero-shot action classification \cite{Ali2024AreVM}. The field has evolved from the use of simple hand-crafted prompts such as “a photo of a \texttt{[y]}” to methods that employ large language models to generate richer and more detailed category descriptions. This shift has produced consistent gains. For instance, CuPL reports improvements of more than one percentage point on ImageNet while remaining fully zero-shot \cite{Cai2025FromLD,Radford2021LearningTV,Pratt2022WhatDA}.
\section{Data and Methods}

\subsection{Dataset}
\begin{figure}[ht]
  \centering
  % Replace with your composed panel image of random samples
  \includegraphics[width=\linewidth]{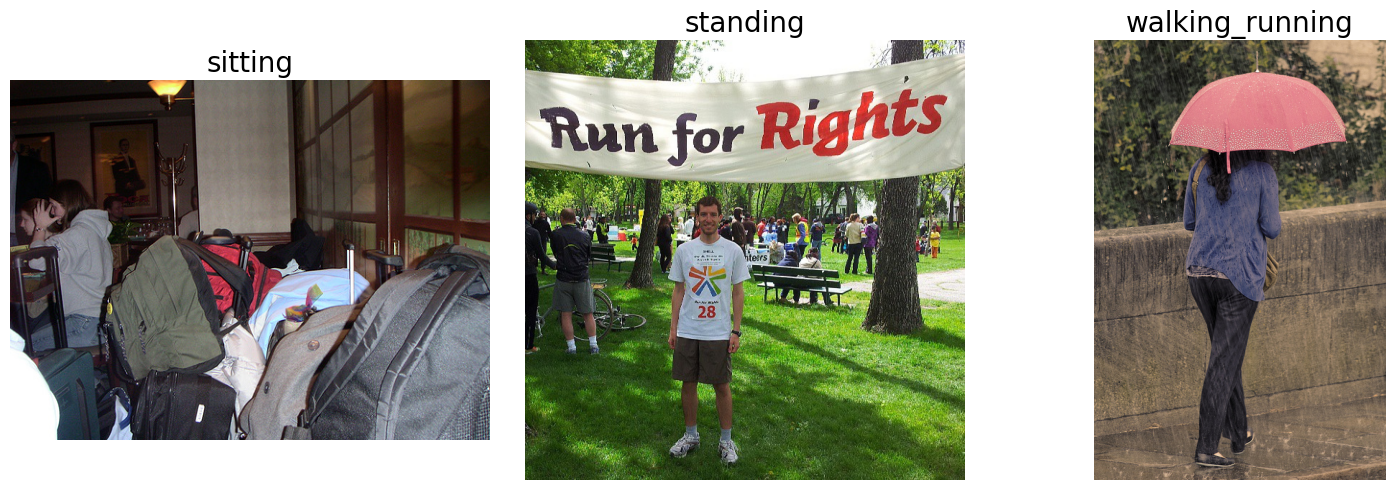}
  \caption{Random samples from the curated MS~COCO subset for \emph{sitting}, \emph{standing}, and \emph{walking or running}.}
  \label{fig:coco_samples}
\end{figure}

\begin{figure}[t]
  \centering
  \includegraphics[width=.78\linewidth,trim=10 10 10 10,clip]{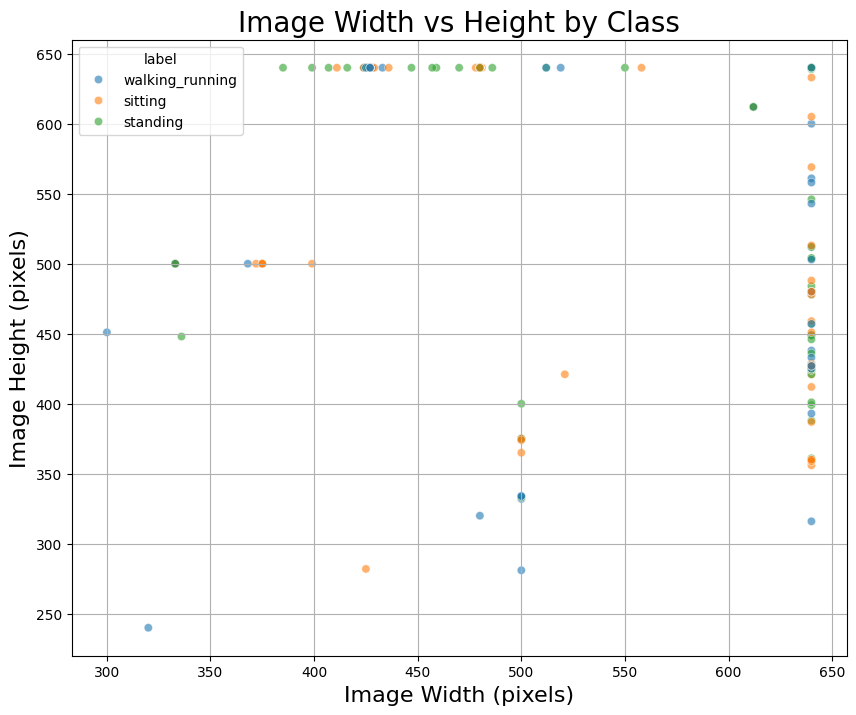}
  \caption{Raw image sizes in the curated subset.}
  \label{fig:image_size}
\end{figure}

\begin{figure}[t]
  \centering
  \includegraphics[width=.78\linewidth,trim=10 10 10 10,clip]{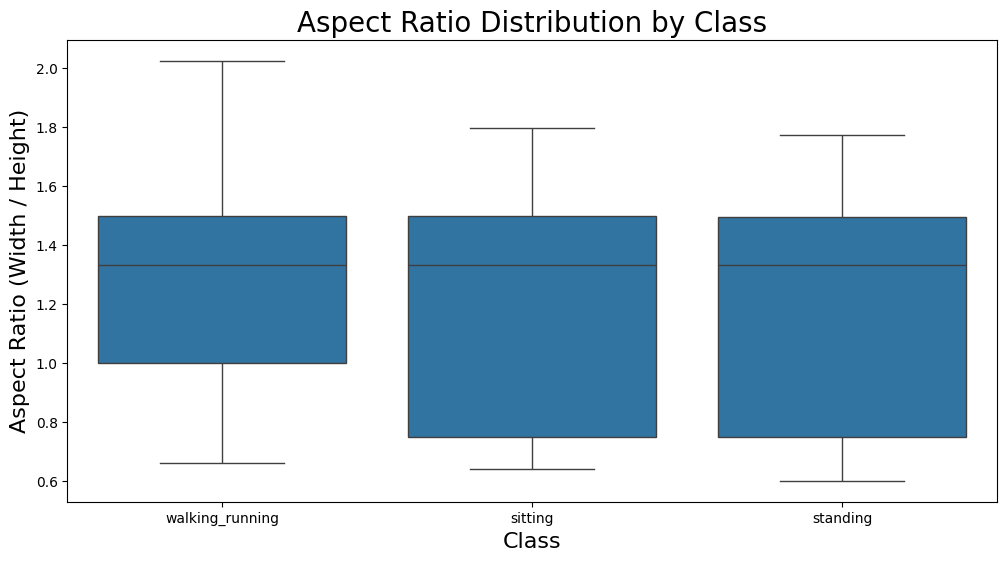}
  \caption{Aspect ratio (width/height) by class.}
  \label{fig:aspect_ratio}
\end{figure}

% We use a curated 285-image subset of the MSCOCO benchmark by \citet{Lin2014MicrosoftCCA}. The dataset is balanced across three action classes: sitting ($n=95$), standing ($n=92$), and walking/running ($n=98$).  Exploratory data analysis confirmed that image dimensions clustered around $640 \times 480$ pixels, justifying a uniform resize to $224 \times 224$ during preprocessing with negligible class-specific distortion.
We evaluate on a curated 285–image subset of MS~COCO \citet{Lin2014MicrosoftCCA}, a large–scale benchmark of everyday scenes with dense instance annotations. From the 2014 releases, images containing at least one visible person with sufficient visual evidence to judge posture were sampled, and a single action label was assigned per image by manual inspection. The subset is balanced across three classes: \emph{sitting} ($n{=}95$), \emph{standing} ($n{=}92$), and \emph{walking or running} ($n{=}98$). Figure~\ref{fig:coco_samples} shows randomly selected examples for each class and illustrates variation in viewpoint, background, and occlusion. Exploratory analysis of raw image sizes in Figure~\ref{fig:image_size} showed a concentration around $640{\times}480$ pixels, which supports a uniform resize to $224{\times}224$ for all models. Aspect–ratio distributions did not differ across classes as seen in Figure~\ref{fig:aspect_ratio}, and visual checks confirmed negligible class–specific distortion after resizing.

\subsection{Models and Experimental Approach}
Our evaluation is structured around three distinct representation paradigms. In each case, the pre-trained model serves as a feature extractor, with a lightweight classifier trained on the resulting embeddings.

\subsubsection{Unimodal Vision Models}
\paragraph{Vision Transformer (ViT)} A pre-trained ViT \citep{Sreekanth_2024} model (\texttt{vit-base-patch16-224}) was fine-tuned on the dataset. It was evaluated on both a binary task (sitting vs. standing) and the full three-class task to assess its generalization capability.

\paragraph{DinoV3} A state-of-the-art vision model pre-trained using self-supervised learning on images alone, allowing us to assess the efficacy of purely visual, non-linguistic representation learning \cite{siméoni2025dinov3}.

\subsubsection{Multimodal Vision-Language Models}
\paragraph{OpenCLIP} We employed OpenAI \cite{Radford2021LearningTV}'s CLIP pre-trained vision encoder \texttt{openai/clip-vit-base-patch32} model, leveraging its extensive image-text pre-training to generate semantically rich embeddings.
\paragraph{Meta CLIP 2} An evolution of CLIP, pre-trained on a more meticulously curated dataset to enhance the quality and robustness of its visual-semantic representations \cite{chuang2025metaclip2worldwide}.
\paragraph{SigLip} A VLM employing a sigmoid-based loss function during pre-training, offering an alternative to the contrastive objective of CLIP \cite{zhai2023sigmoidlosslanguageimage}.

\subsubsection{Pose-Centric Structural Model}
\paragraph{YOLOv11x-pose} This model implements a two-stage process. First, the YOLOv11x pose \cite{khanam2024yolov11overviewkeyarchitectural} architecture is applied to each image to extract a set of 2D keypoints that represents the subject's skeleton. Second, geometric features, such as the angles between the left and right knee and hip joints, are calculated from these key points. A simple classifier is then trained on these angular features to determine the final action class.

\subsection{Prompt tiering for zero-shot classification}
We vary only the specificity of the text prompt in order to test how wording affects zero-shot posture recognition with scarce data. \textbf{Tier 1} uses the class label in a minimal template such as “a photo of a person \texttt{[class]}”, which reflects common zero-shot practice. \textbf{Tier 2} adds a brief action cue that clarifies the target category, for example “a person seated on a chair”, “a person standing still and upright”, or “a person mid-stride with one foot off the ground”. \textbf{Tier 3} replaces action words with short anatomical or pose constraints, for example “hips and knees bent at right angles” for sitting or “legs straight and torso vertical” for standing. Across tiers we keep prompts scene free and we avoid background, clothing, and identity terms so that only pose information differs. For each tier we create one prompt per class, compute unit-normalized text embeddings once, embed each image once at the model’s native resolution, and score classes by cosine similarity. We report accuracy and macro-F1 per tier on the same images and preprocessing settings without any model fine-tuning so that observed differences can be attributed to prompt content rather than changes in data or optimization.

\subsection{Experimental Approach}

All experiments were conducted with a focus on reproducibility and were run on Google Colab on a single NVIDIA T4 GPU with 16GB of memory. The dataset was partitioned on a fixed stratified 80\% training, 10\% validation, and 10\% test split for all experiments. A global random seed was established to ensure that all models were trained and evaluated on the exact same data partitions.

The task was defined in two distinct classification scenarios to assess the performance of the model at varying levels of difficulty: (1) A simplified binary task focusing on two more visually distinct classes: \textit{sitting} vs. \textit{walking/running}. (2) Three-class task encompassing all labels: \textit{sitting}, \textit{standing}, and \textit{walking/running}; To account for stochasticity in the training process, each model was trained and evaluated over five independent runs with different seeds. We used an early stopping mechanism with patience of 5 epochs, monitoring the validation loss to prevent over-fitting. 
\section{Results}

The empirical evaluation is presented in two parts. (1) A comparative analysis of different model architectures under a standard training and evaluation paradigm to establish baseline performance. (2) A zero-shot experiment investigating how the specificity of text prompts affects the performance of Vision-Language Models.

\subsection{Comparative Analysis of Model Architectures}
A comparative evaluation of models from three different paradigms (Unimodal, VLM, and Pose-Centric) was conducted across both binary and multi-class classification tasks as the first stage of the evaluation.

\begin{figure}[htbp]
    \centering
    \includegraphics[width=0.45\textwidth]{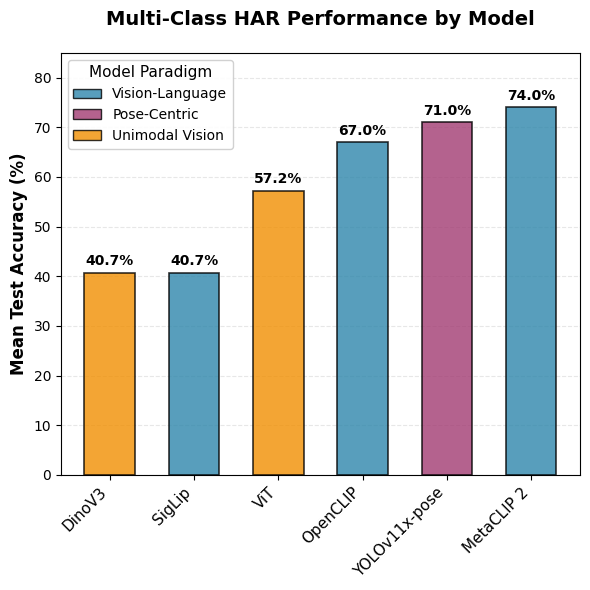}
    \caption{Mean test accuracy on the primary three-class HAR task. Models with semantic (VLM) or structural (Pose) priors demonstrate a clear performance advantage.}
    \label{fig:multiclass_accuracy}
\end{figure}

\subsubsection{Performance on Binary Classification}

To investigate how model performance is affected by task complexity, the models were evaluated on a binary classification task (sitting vs. walking/running). The results, presented in Table~\ref{tab:model_performance}, show a general performance uplift across most models, yet the relative ranking remains largely consistent.

MetaCLIP 2 performed the best with an accuracy of 92.8\%. Notably, ViT performed well in this less ambiguous setting, achieving 90.0\% accuracy, nearly matching the top VLM. This suggests that when classes are more visually distinct, a powerful unimodal architecture can be highly effective. The original OpenCLIP model also performed strongly at 88.1\%. However, DinoV3 and SigLip surprisingly continued to lag significantly, with accuracies of 57.5\% and 56.5\%, respectively.

\subsubsection{Performance on Multi-Class Classification}

The three-class action recognition task represents the core challenge of this study, requiring models to distinguish between visually similar and ambiguous postures from a single static frame. As detailed in Figure~\ref{fig:multiclass_accuracy}, the performance of the evaluated models diverged significantly, clearly separating them into distinct tiers.

The models endowed with strong priors formed the top tier. The Vision-Language Model MetaCLIP 2 achieved the highest accuracy at 74.0\%. Following closely was the YOLOv11x-pose model, which, by leveraging a structural representation of the human body, secured an accuracy of 71.0\%. The original OpenCLIP model also delivered a robust performance of 67.0\%.

A substantial performance gap exists between these models and the unimodal models that learn from pixels alone. The standard Vision Transformer (ViT) achieved a modest accuracy of 57.2\%. The purely self-supervised DinoV3 and the VLM SigLip both struggled significantly, each attaining only 40.7\% accuracy, a result only marginally better than random chance.

\begin{table}[ht]
\centering
\resizebox{\linewidth}{!}{%
\begin{tabular}{lccccc}
\hline
\textbf{Model} & \textbf{B. Acc.} & \textbf{M. Acc.} & \textbf{Prec.} & \textbf{Rec.} & \textbf{F1} \\
\hline
MetaCLIP 2      & 0.92 & 0.74 & 0.74 & 0.74 & 0.74 \\
ViT             & 0.90 & 0.52 & 0.59 & 0.57 & 0.57 \\
OpenCLIP            & 0.88 & 0.67 & 0.68 & 0.67 & 0.66 \\
YOLOv11x-pose   & ---  & 0.71 & 0.73 & 0.71 & 0.71 \\
DinoV3          & 0.57 & 0.40 & 0.41 & 0.41 & 0.40 \\
SigLip         & 0.56 & 0.40 & 0.28 & 0.41 & 0.33 \\
\hline
\end{tabular}%
}
\caption{
Performance comparison of vision models, reporting \textbf{(B. Acc.)} Binary Accuracy, \textbf{(M. Acc.)} Multi-class Accuracy, \textbf{(Prec.)} Macro Precision, \textbf{(Rec.)} Macro Recall, and \textbf{(F1)} Macro F1 Score.
}
\label{tab:model_performance}
\end{table}

\subsubsection{Analysis of Class-Specific Metrics}

To gain a more nuanced understanding, we analyzed the macro-averaged Precision, Recall, and F1-Score (Table~\ref{tab:model_performance}). These metrics reinforce the hierarchy observed in accuracy. MetaCLIP 2 and YOLOv11x-pose demonstrated a strong balance between precision and recall, resulting in high F1-Scores of 0.74 and 0.71, respectively, indicating reliable classification across all three categories. In contrast, lower-performing models exhibited imbalances. For instance, SigLip2 had a recall of 0.41 but a very low precision of 0.28, suggesting it generated a large number of false positive predictions in its attempt to classify instances from all classes.

\subsection{Prompt-Specific Zero-Shot Performance}
In our second set of experiments, we investigated how prompt specificity affects the zero-shot performance of VLMs. The results, detailed in Table~\ref{tab:combined_classification_tiers}, show that the relationship between prompt detail and model performance is not linear and is highly model-dependent.

\subsubsection{Performance Trends for High Performing Models}
The primary trend observed for the leading Vision-Language Models is a clear inverse relationship between prompt specificity and classification performance. As represented in Table~\ref{tab:combined_classification_tiers} both MetaCLIP 2 and OpenCLIP, the simplest Tier 1 prompts consistently achieved the highest accuracy and F1 scores. The introduction of more descriptive features in Tier 2 or anatomical cues in Tier 3 resulted in a significant degradation of performance. This effect was particularly pronounced for MetaCLIP 2, where the multi-class accuracy fell sharply from 68.8\% with a Tier 1 prompt to 55.1\% with a Tier 2 prompt. Similarly, OpenCLIP's multi-class accuracy saw a substantial decrease from a high of 71.2\% (Tier 1) to 52.6\% (Tier 2). This consistent impact suggests a phenomenon of ``prompt overfitting'' where excessive detail may unduly constrain the models and hinder their ability to generalize.

\subsubsection{Model-Dependent Responses to Prompt Granularity}
The Tier 1 performance trend was not universal, highlighting that the optimal prompt strategy is highly model dependent. The lower-performing SigLip model exhibited a contrasting response to the increase in prompt detail. While its overall accuracy remained consistently low, its ability to classify the ambiguous \texttt{walking\_running} class was significantly boosted by the specific, ``body cue-based'' Tier 3 prompts. This was most evident in the binary task, where the F1 score for this specific class jumped from 0.364 with a basic Tier 1 prompt to 0.566 with the detailed Tier 3 prompt.

\begin{table*}[ht]
\centering
\begin{tabular}{llcccccc}
\hline
\textbf{Task} & \textbf{Tier} & \multicolumn{2}{c}{\textbf{MetaCLIP 2}} & \multicolumn{2}{c}{\textbf{OpenCLIP}} & \multicolumn{2}{c}{\textbf{SigLIP}} \\
\cline{3-8}
 & & \textbf{Accuracy} & \textbf{Macro F1} & \textbf{Accuracy} & \textbf{Macro F1} & \textbf{Accuracy} & \textbf{Macro F1} \\
\hline
\multirow{3}{*}{\textbf{Binary}} & Tier 1 & 0.938 & 0.938 & 0.907 & 0.907 & 0.565 & 0.516 \\
 & Tier 2 & 0.751 & 0.742 & 0.850 & 0.847 & 0.523 & 0.490 \\
 & Tier 3 & 0.731 & 0.714 & 0.876 & 0.875 & 0.539 & 0.537 \\
\hline
\multirow{3}{*}{\textbf{Multi-class}} & Tier 1 & 0.688 & 0.686 & 0.712 & 0.708 & 0.365 & 0.346 \\
 & Tier 2 & 0.551 & 0.508 & 0.526 & 0.533 & 0.316 & 0.259 \\
 & Tier 3 & 0.565 & 0.528 & 0.628 & 0.628 & 0.312 & 0.302 \\
\hline
\end{tabular}
\caption{Classification Performance (Binary and Multi-class) across different tiers of prompts}
\label{tab:combined_classification_tiers}
\end{table*}
\section{Discussion \& Conclusion}

\begin{figure*}[t]
    \centering
    \begin{subfigure}[b]{0.70\linewidth}
        \centering
        \includegraphics[width=\linewidth]{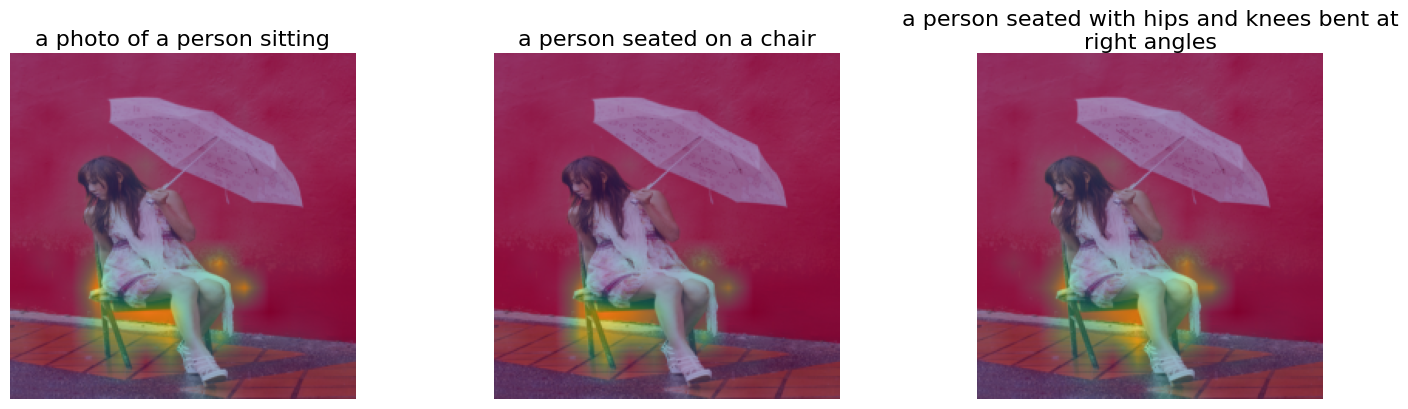}
        \caption{Grad-CAM for three phrasings of the “sitting” concept, showing saliency on the chair and hip–knee region.}
        \label{fig:gradcam_sitting}
    \end{subfigure}
    \hfill
    \begin{subfigure}[b]{0.70\linewidth}
        \centering
        \includegraphics[width=\linewidth]{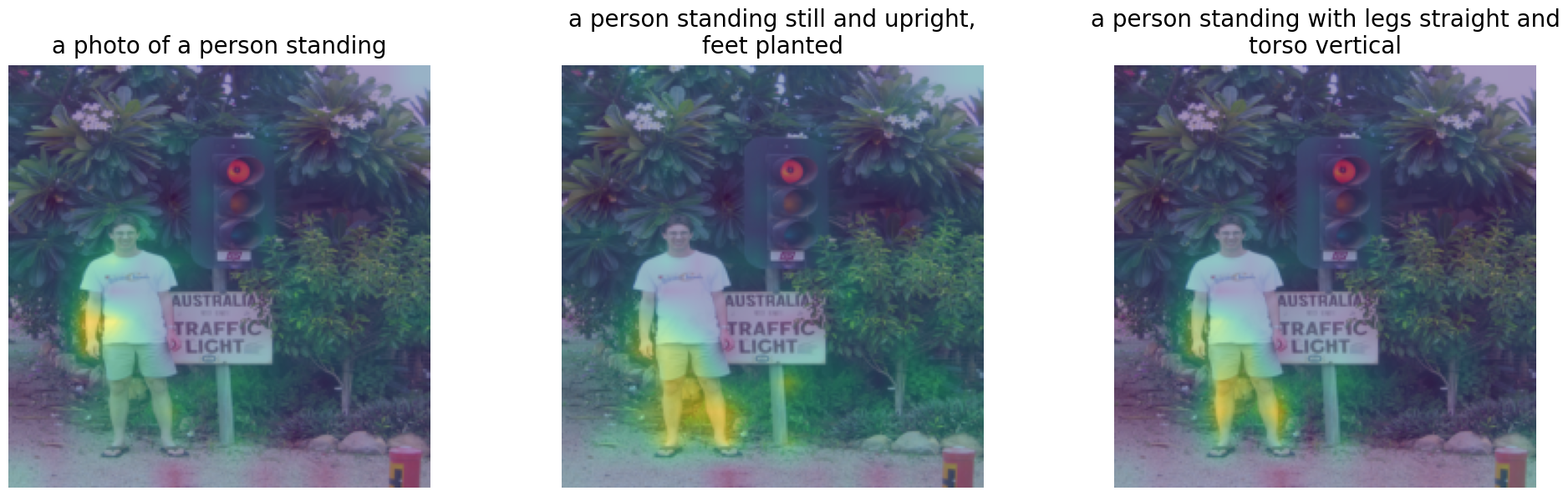}
        \caption{Grad-CAM for three phrasings of the “standing” concept, showing saliency on the legs and torso.}
        \label{fig:gradcam_standing}
    \end{subfigure}
    \caption{
    Grad-CAM visualizations for different phrasings of the concepts “sitting” and “standing”. Increased specificity in phrasing leads to more focused saliency in relevant body regions.
    }
    \label{fig:gradcam_concepts}
\end{figure*}

\subsection{Prompt Specificity as Supervision at Inference}
Prompt wording functions as an explicit prior on the classifier decision in a zero-shot setting. Minimal, noun-centric prompts align with the distributions seen during pre-training of image–text encoders, where concept names are frequent and broadly grounded. This alignment explains the competitiveness of label-only prompts in closed sets. In contrast, adding brief action cues can introduce a linguistic–visual mismatch for still images, since verbs such as “walking” or “standing still” denote dynamics or intent rather than stable appearance. The resulting text embeddings are drawn toward contexts that are weakly supported by a single frame, which reduces similarity margins and increases overlap between neighbouring classes.

Geometric phrasing exerts a different influence. Short anatomical constraints specify local, view-stable relations, for example relative angles at the hip and knee or verticality of the torso, that are directly verifiable in a single image. Gradient-based attributions (see Figure~\ref{fig:gradcam_sitting}) consistently show increased concentration over limb and torso regions when such constraints are used, and decreased reliance on background texture or incidental objects. The benefit is class dependent. Categories that are well captured by a nominal phrase, such as sitting in uncluttered scenes, receive limited additional gain. Categories that are visually adjacent in a still image, such as standing versus walking or running, benefit from geometric prompts because these encode spatial structure that separates the classes without introducing scene bias.

These observations support a simple policy for low-resource use. Prefer label-style prompts as the default in closed-set classification with pre-trained encoders. Introduce compact geometric descriptors selectively for pairs that remain ambiguous, and verify with attribution that attention shifts from background to pose-relevant regions. Reserve action verbs for cases where the class definition truly requires dynamic semantics, since such wording is not consistently grounded in single images.

\subsection{Comparative Model Behaviour and Calibration}
Across encoders, closed-set zero-shot performance tracks the pre-training objective. OpenCLIP and MetaCLIP optimise a soft-max contrastive loss with a learned temperature, which induces competition among text candidates and yields larger similarity margins in classification. SigLIP optimises independent sigmoid scores for pairs, which favours retrieval but produces flatter score distributions in a closed set. The flatter distributions manifest as smaller top-1 minus top-2 margins and greater sensitivity to prompt phrasing, particularly for visually adjacent classes.

Calibration follows the same pattern. After unit normalisation of embeddings, a single temperature applied to cosine similarities brings CLIP-family confidences into closer agreement with accuracy. The same treatment is less effective for SigLIP model because the training objective does not enforce cross-class competition, and confidence therefore reflects pairwise affinity rather than calibrated class probability. Reliability curves and expected calibration error consequently favour OpenCLIP and MetaCLIP 2 under a shared temperature, whereas SigLIP remains comparatively miscalibrated or require tier-specific scaling.

Baselines clarify the role of alignment and structure. DINOv3 and a standard ViT combined with frozen sentence embeddings underperform and calibrate poorly because the image and text spaces are learned independently rather than jointly. YOLOv11-Pose with a simple geometric decision rule is competitive when keypoints are detected with confidence, which indicates that explicit pose structure can substitute for language supervision when the detector is reliable. Taken together, these observations suggest that, in data-scarce image-based recognition, cross-modal alignment with a contrastive objective provides stronger closed-set behaviour, while geometric priors provide a complementary path when alignment is weak or text supervision is constrained.

\subsection{Language-Free and Pose-Based Baselines under Data Scarcity}
Vision-only encoders such as DINOv3 and a standard ViT provide a language-free reference that isolates the value of cross-modal alignment. When image embeddings are compared to sentence embeddings from an unrelated text model, the spaces are not jointly learned. As a result the cosine geometry reflects two independent objectives rather than class evidence. This mismatch explains the weaker separability and the poor calibration that appear even when preprocessing is held constant. The baselines are therefore informative as a lower bound. They confirm that generic visual features carry some signal for posture, yet they also show that alignment with text during pre-training is the primary driver of robust zero-shot classification.

A pose-based baseline introduces a different kind of supervision that is structural rather than linguistic. YOLOv11-Pose produces 2D keypoints, and a deterministic rule maps joint configuration to the three classes. When detections are confident, the rules are competitive because they test explicit geometric relations that are stable in a single frame. However, performance depends on detection coverage. Occlusion, truncation, unusual viewpoints, and multiple persons reduce keypoint quality and lead to abstentions or incorrect geometry, which directly lowers accuracy. Reporting coverage alongside accuracy is therefore necessary. On the covered subset the baseline demonstrates that posture can be resolved without any text, while the uncovered subset clarifies where structural priors fail.

These baselines contribute two practical insights for low-resource use. First, if language supervision is restricted because of privacy or deployment constraints, a pose pipeline can recover a substantial fraction of performance provided that person detection is dependable. Second, if a language-free heuristic is required for simplicity, cosine scoring between DINOv3 or ViT features and frozen sentence embeddings should be treated as a diagnostic tool rather than as a calibrated classifier. In combination with the multimodal results, the baselines indicate that cross-modal alignment should be the default, and that explicit pose structure is a useful fallback when alignment is unavailable or when prompts cannot be used.

\subsection{Attention Maps and Error Patterns in Still-Image HAR}
Attribution on the sitting examples shows a consistent shift as prompt specificity increases. The label-style prompt yields broad responses that cover the person and nearby objects. Adding an action cue narrows the response toward the pelvis and the supporting surface. Geometric phrasing concentrates the map on hips, knees, and the contact region with the chair. This progression indicates that geometric wording encourages the model to prefer pose evidence over contextual cues.

For standing, the label-style prompt again produces diffuse maps with noticeable activation on salient background regions as seen in Figure~\ref{fig:gradcam_standing}. The action cue that mentions stillness reduces spread and increases activation around the legs and feet. The geometric formulation further localises energy along the vertical axis of the body, especially the shins and torso. When predictions are incorrect for standing, the maps typically remain broad and include background structure, which suggests insufficient reliance on limb configuration in those cases.

Two practical uses follow. First, attribution can serve as a prompt diagnostic: adopt geometric phrasing when maps remain diffuse under a label-style prompt, and retain the minimal prompt when maps are already concentrated on limbs and joints. Second, report simple map statistics alongside accuracy, such as the proportion of normalised heat inside a person region and the entropy of the map. Higher in-person proportion and lower entropy correlate with the tighter, pose-focused responses observed for the geometric prompts in both sitting and standing.

\subsection{Practical Implications, Robustness, and Limitations}
\paragraph{Practical implications} In data-scarce settings, a label-style prompt for each class with unit-normalised embeddings and a single temperature applied to cosine scores is a strong baseline. When confusions remain for visually adjacent categories, replace the label with a compact geometric description for those specific classes. Monitor decision confidence with the top-one minus top-two similarity margin. Introduce an abstention rule based on a margin threshold for low-confidence cases. If language supervision is not available, a pose pipeline that uses YOLOv11-Pose with a deterministic geometric rule provides an alternative, provided that keypoint detection is reliable.

\paragraph{Robustness considerations} Performance depends on image framing and resolution. Crops that remove feet or hips reduce margins for posture classes, therefore detection and resizing should preserve the lower body. Resolution influences prompts that encode limb configuration. The native 224 input supports fair comparison, while higher resolution can improve separation when resources allow. Paraphrases within a tier can shift scores, so a small prompt ensemble per class stabilises predictions with limited overhead. Calibration differs across encoders. Fit a single temperature once per model and keep it fixed across tiers to preserve comparability. For the pose baseline, report coverage since occlusion, truncation, and small subjects reduce the fraction of usable detections.

\paragraph{Limitations} The study uses a small COCO-derived subset with three classes and single-person images, which narrows external validity. Only still images are considered, therefore temporal cues are absent. Prompts are written in English and the evaluation covers a limited set of encoders and one attribution method. The conclusions apply to image-based posture recognition under severe data scarcity and should be revisited with larger datasets, multilingual prompts, multi-person scenes, and alternative backbones and attribution methods.

\section{Conclusion}
This study examined zero-shot posture recognition from still images under data scarcity by treating language as the label at inference time. Across OpenCLIP, MetaCLIP 2, SigLIP families, vision-only baselines with DINOv3 and ViT, and a pose-based baseline with YOLOv11-Pose, results show that prompt wording acts as an explicit prior. Label-style prompts provide strong closed-set performance, while compact geometric phrasing improves separability for visually adjacent classes when margins are small. Action cues that describe dynamics offer limited benefit for single images. Calibration and margin analysis favour contrastive vision–language encoders, and the pose pipeline serves as a viable alternative when language supervision is restricted. The findings yield a simple policy for deployment in low-resource settings and motivate extensions to larger class sets, multilingual prompts, multi-person scenes, and additional backbones and datasets.

\bibliography{custom}

\appendix

\end{document}